\documentclass[shortpaper]{clv3}

\usepackage{savesym}
\usepackage{xcolor}
\usepackage{times,graphicx}
\usepackage{latexsym,booktabs} %,dingbat}
\usepackage[T1,T5]{fontenc}
\usepackage[utf8]{inputenc}
\usepackage{amsmath}
\savesymbol{numdef}
\usepackage{multirow}
\usepackage{multicol,float,graphicx}
\usepackage{url}
\usepackage{natbib}
\usepackage{tikz}
\usepackage{tikz-dependency}
\usepackage{collcell}
\usepackage{todonotes}
\usepackage{colortbl,dcolumn}
\usepackage{hyperref}
\usepackage{cleveref}
\makeatletter
\crefformat{section}{\S#2#1#3} % see manual of cleveref, section 8.2.1
\crefformat{subsection}{\S#2#1#3}
\crefformat{subsubsection}{\S#2#1#3}

\definecolor{darkblue}{rgb}{0, 0, 0.5}
\hypersetup{colorlinks=true,citecolor=darkblue, linkcolor=darkblue, urlcolor=darkblue}
\runningtitle{What Should/Do/Can LSTMs Learn When Parsing Auxiliary Verb Constructions?}
\runningauthor{de Lhoneux et al.}
\historydates{
  Submission received:            18 July 2019;
  revised version received:      19 May 2020;
  accepted for publication:       3 October 2020.
  }

\newtoggle{inTableHeader}% Track if still in header of table
\toggletrue{inTableHeader}% Set initial value
\newcommand*{\StartTableHeader}{\global\toggletrue{inTableHeader}}%
\newcommand*{\EndTableHeader}{\global\togglefalse{inTableHeader}}%
\let\OldTabular\tabular%
\let\OldEndTabular\endtabular%
\renewenvironment{tabular}{\StartTableHeader\OldTabular}{\OldEndTabular\StartTableHeader}%

\newcommand*{\MinNumber}{-20}%
\newcommand*{\MidNumber}{0.0} %
\newcommand*{\MaxNumber}{20}%

\newcommand{\ApplyGradient}[1]{%
    \ifdim #1 pt > \MidNumber pt
    \pgfmathsetmacro{\PercentColor}{max(min(100.0*(#1 - \MidNumber)/(\MaxNumber-\MidNumber),100.0),0.00)} %
    \hspace{-0.33em}\colorbox{green!\PercentColor!yellow}{\makebox[2em]{#1}}
    \else
    \pgfmathsetmacro{\PercentColor}{max(min(100.0*(\MidNumber - #1)/(\MidNumber-\MinNumber),100.0),0.00)} %
    \hspace{-0.33em}\colorbox{red!\PercentColor!yellow}{\makebox[2em]{#1}}
    \fi
}

\newcommand*{\ApplyGradientB}[1]{\iftoggle{inTableHeader}{#1}{\ApplyGradient{#1}}}%
\newcolumntype{R}{>{\collectcell\ApplyGradientB}c<{\endcollectcell}}

\title{What Should/Do/Can LSTMs Learn \\When Parsing Auxiliary Verb Constructions?}

\author{Miryam de Lhoneux
\thanks{Work carried out while at Uppsala University.}
}
\affil{ Department of Computer Science\\
University of Copenhagen\\
}
\author{Sara Stymne}
\affil{ Department of Linguistics and Philology \\
Uppsala University }
\author{Joakim Nivre}
\affil{ Department of Linguistics and Philology \\
Uppsala University }

\date{}

\begin{document}
\maketitle

\begin{abstract}
    There is a growing interest in investigating what neural NLP models learn about language.
    A prominent open question is the question of whether or not it is necessary to model hierarchical structure. 
    We present a linguistic investigation of a neural parser adding insights to this question.
    We look at transitivity and agreement information of auxiliary verb constructions (AVCs) in comparison to finite main verbs (FMVs). This comparison is motivated by theoretical work in dependency grammar and in particular the work of \citet{tesniere59} where AVCs and FMVs are both instances of a nucleus, the basic unit of syntax. An AVC is a dissociated nucleus, it consists of at least two words, and an FMV is its non-dissociated counterpart, consisting of exactly one word. We suggest that the representation of AVCs and FMVs should capture similar information.
    We use diagnostic classifiers to probe agreement and transitivity information in vectors learned by a transition-based neural parser in four typologically different languages. We find that the parser learns different information about AVCs and FMVs if only sequential models (BiLSTMs) are used in the architecture but similar information when a recursive layer is used. We find explanations for why this is the case by looking closely at how information is learned in the network and looking at what happens with different dependency representations of AVCs. We conclude that there may be benefits to using a recursive layer in dependency parsing and that we have not yet found the best way to integrate it in our parsers.
\end{abstract}

\section{Introduction}
\indent In the past few years, the interest in investigating what neural models learn about language has been growing. This can be interesting from both a machine learning perspective, to better understand how our models work, and from a linguistic perspective, to find out what aspects of linguistic theories are important to model. 
A popular method is the use of \emph{diagnostic classifiers} \citep{hupkes2018visualisation}
where the idea is to probe whether or not a model trained for a source task (for example language modelling) learns a target task (for example subject verb agreement) as a byproduct of learning the source task. This is done by training vectors on the source task, freezing these vectors and using them to train a classifier on a target task. If we can successfully train that classifier, we have some indication that the target task has been learned as a byproduct of learning the source task.
In this article, we use this method to investigate whether or not a specific aspect of linguistic theory is learned, the notion of dissociated nucleus from dependency grammar (explained in \cref{sec:avc}), as a byproduct of learning the task of dependency parsing for several languages. For this, we focus on auxiliary verb constructions (AVCs).

A prominent question in neural modelling of syntax is the question of whether or not it is necessary to model hierarchical structure. Sequential models (long short-term memory networks (LSTMs)) have shown surprising capabilities at learning syntactic tasks \citep{linzen16assessing,gulordava18colorless}, and models of dependency parsing using sequential models are very accurate \citep{kiperwasser16}. While recursive neural networks surpass the abilities of sequential models for learning syntactic tasks \citep{kuncoro2018lstms}, their use in dependency parsing seems superfluous compared to using sequential models when looking at parsing accuracy \citep{delhoneux19recursive}. However, there may be benefits to using recursive neural networks in parsing that are not reflected in parsing accuracy.
In particular, for reasons outlined in \cref{sec:recnns}, they may be useful when it comes to learning the notion of dissociated nucleus. In this article, we evaluate the usefulness of recursive neural networks when it comes to learning the notion of dissociated nucleus.

The goals of this article are thus threefold. First, we look at dependency grammar to theoretically motivate that our models \emph{should} learn the notion of dissociated nucleus. Second, we develop a method using diagnostic classifiers to test whether or not our models \emph{do} learn this notion. Third, we investigate the role that recursive neural networks play in learning this notion and test whether or not our models \emph{can} learn this notion when augmented with a recursive layer.

\section{Background and Research Questions}
\subsection{AVCs in Dependency Grammar and Dependency Parsing}
\label{sec:avc}
Dependency parsing has gained popularity in the last 15 years, drawing ideas from dependency grammar but diverging from it in important ways. Research on dependency parsing has relied on a definition of dependency trees where the basic units of syntax are words and the relations that hold between words in a sentence are binary asymmetric relations. In most dependency grammar theories, representations are considerably more complex, often consisting of multiple strata as in Meaning-Text Theory \citep{melcuk88} and Functional Generative Description \citep{sgall86}. In the seminal work of \citet{tesniere59}, a single level of representation is used but the  basic unit of syntax is not the word but the more abstract notion of \emph{nucleus}. Nuclei often correspond to individual words but sometimes correspond to several words, typically a content word together with one or more function words, which are said to constitute a \emph{dissociated nucleus}. The internal elements of a dissociated nucleus are connected by \emph{transfer} relations, while the nuclei themselves are connected by \emph{dependency} relations or (in the case of coordination) by \emph{junction} relations.

In the dependency parsing literature, a sentence with an auxiliary is usually represented as either the top or the bottom tree in the left part of Figure~\ref{fig:nucl}, with either the auxiliary or the main verb being dependent on the other. If we follow the ideas from \citet{tesniere59}, it can be represented as in the right part of Figure~\ref{fig:nucl} where the auxiliary and main verb are connected by a transfer relation to form a nucleus. This nucleus is itself connected to the other words/nuclei in the sentence. In this example, the word \emph{that} corresponds to a nucleus and the words \emph{could} and \emph{work} are each part of a dissociated nucleus.
In this sense, the definition of dependency trees that is used in dependency parsing is a simplification compared to the representations used in Tesni\`{e}re's dependency grammar. We are losing the information that the relation between \emph{could} and \emph{work} is a different type of relation than the relation between \emph{that} and \emph{work}.\footnote{Note that the labels disambiguate these cases but there are more labels than these two and labels do not encode information about whether they are dependency or transfer relations.} We are forced to choose a head between the main verb and the auxiliary, even though the main verb and auxiliary share head properties: inflectional verbal features like agreement, tense, aspect, mood, etc. are typically encoded in the auxiliary whereas lexical features like valency are properties of the main verb.
As \citet{williams2018latent} have shown, a network that learns latent trees as part of a downstream task does not necessarily learn trees that correspond to our linguistic intuitions. This means that teaching our models the difference between transfer and dependency relations will not necessarily make a difference when it comes to downstream tasks. However, it is still informative to find out if we can learn a representation that is linguistically motivated, to understand better how our models represent certain linguistic phenomena.
We can subsequently investigate if learning this type of representation is useful for downstream tasks which we leave to future work. In turn, this could also inform linguistic theory, by finding if this notion is relevant or not for practical language technology.

\begin{figure}
    \centering
    \begin{dependency}[theme=simple]
      \begin{deptext}[column sep=0.8em]
            that \& could \& work\\
        \end{deptext}
        \depedge{3}{1}{}
        \depedge{3}{2}{}
        \depedge[edge below]{2}{1}{}
        \depedge[edge below]{2}{3}{}

    \end{dependency}
    \quad 
    \begin{dependency}[theme=simple]
        \begin{deptext}[column sep=.8em]
            that \& could \& work\\
            \textcolor{white}{.}\\
        \end{deptext}
        \wordgroup{1}{2}{3}{dn}
        \wordgroup{1}{1}{1}{sn}
        \groupedge{dn}{sn}{}{0}
    \draw [thick, blue] (\wordref{1}{2})--(\wordref{1}{3});
    \end{dependency}
    \label{fig:nucl}
    \caption{Two different representations of a sentence with auxiliary as used in dependency parsing (left) vs as can be represented following \protect{\citet{tesniere59}} (right).}
\end{figure}

\indent Universal Dependencies (UD) \citep{ud} is a project that is seeking to harmonise the annotation of dependency treebanks across languages. Having such harmonised annotations makes it easier to incorporate linguistic information into parsing models while remaining language independent. In particular, as described by \citet{nivre15cicling}, UD adopts an analysis of language where function words attach to content words. This analysis, he argues, can be interpreted as a dissociated nucleus, as defined by \citet{tesniere59}. However, this notion has not been made explicit when training parsers.
In pre-neural transition-based parsers like \citet{nivre06malt}, when a dependent gets attached to its head, features of the head are still used for further parsing but features of the dependent are usually discarded.\footnote{Although features of the dependent can be used as features of the head.}

In neural parsers, it is less clear what information is used by parsers. Current state-of-the-art models use (Bi)LSTMs \citep{dyer15,kiperwasser16,dozat16biaffine}, and LSTMs make it possible to encode information about the surrounding context of words in an unbounded window (which is usually limited to a sentence in practice). 
In this article, we take a step in finding out what neural parsers learn by testing if they capture the notion of dissociated nuclei. We do this by looking in detail at what a BiLSTM-based parser learns about a specific type of dissociated nucleus: auxiliary verb constructions. %\\
We focus on AVCs as they are a typical example of dissociated nucleus and are well attested typologically, see for example \citet{anderson2011auxiliary}. We focus on 4 different languages, for reasons explained in \cref{sec:exp}. 

We can use diagnostic classifiers to look at whether information like valency, agreement, tense, mood, etc.\ is encoded in a vector representing a word or a subtree in parsing. This allows us to specify how to test whether or not our parser learns the notion of dissociated nucleus. We ask the following question: when making a parsing decision about a nucleus, does the parser have access to similar information regardless of whether the nucleus is dissociated or not? 
For the case of AVCs, this means that the parser should learn similar information about AVCs as it does about their non-dissociated counterpart: simple finite main verbs (henceforth FMV). An example FMV is \emph{did} in the sentence \emph{I did that.}%\\

\subsection{Recursive vs Recurrent Neural Networks}
\label{sec:recnns}
\indent LSTMs are sequential models and therefore do not explicitly model hierarchical structure. \citet{dyer15} have shown that using a recursive layer on top of an LSTM is useful when learning a parsing model. This recursive layer is used to compose the representation of subtrees. 
However, \citet{kiperwasser16} have more recently obtained parsing results on par with the results from \citet{dyer15} using only BiLSTMs. Additionally, recent work has claimed that LSTMs are capable of learning hierarchical structure \citep{linzen16assessing,enguehard17exploring,gulordava18colorless, blevins18deep}. 
This indicates that a \mbox{BiLSTM} might be sufficient to capture the hierarchical structure necessary in parsing. 
However, \citet{kuncoro2018lstms} have also shown that although sequential LSTMs can learn syntactic information, a recursive neural network which explicitly models hierarchy (the Recurrent Neural Network Grammar (RNNG) model from \citet{dyer15}) is better at this: it performs better on the number agreement task from \citet{linzen16assessing}. In addition, \citet{ravfogel2018can} and \citet{ravfogel2019studying} have cast some doubts on the results by \citet{linzen16assessing} and \citet{gulordava18colorless} by looking at Basque and synthetic languages with different word orders respectively in the two studies.\\
\indent Motivated by these findings, we recently investigated the impact of adding a recursive layer on top of a BiLSTM-based parser in \citet{delhoneux19recursive} and found that this recursive layer is superfluous in that parsing model when we look at parsing accuracy.
This indicates that BiLSTM parsers capture information about subtrees, but it is also possible that the advantages and disadvantages of using a recursive layer cancel each other out in the context of our parser and that the advantages of using it are not reflected in parsing accuracy. A recursive layer might still be useful when it comes to learning the notion of dissociated nucleus. As a matter of fact, it might make sense to use recursive composition to model relations of \emph{transfer} and not relations of \emph{dependency} in the sense of \citet{tesniere59}.

\subsection{Research questions}
\label{sec:questions}
\indent We use diagnostic classifiers to probe what information is encoded in vectors representing AVCs learned by a dependency parser. We are interested in finding out whether or not a parser learns similar information about AVCs as they learn about their non-dissociated counterpart, FMVs. In other words, if a parser learns the notion of dissociated nucleus, we expect it to have information about agreement, tense, aspect, mood as well as valency encoded in subtree representations of AVCs to the same extent as it is encoded in vectors representing FMVs. \\
\indent With UD treebanks, it is straightforward to design tasks that probe transitivity in AVCs and FMVs: we can look at objects and indirect objects of the main verb. It is also straightforward to design tasks that probe agreement of AVCs and FMVs: we can use the morphological features that encode information about the subject's number and person. It is less straightforward to design tasks that probe information about tense, mood, and aspect (TMA) because that would require annotation of the verb phrases, since the morphological features of individual verbs do not give enough information. For example, in the AVC \emph{has been}, the tense feature for \emph{has} is \emph{present} and for \emph{been} it is \emph{past} but no feature indicates that the AVC is in the present perfect.\footnote{See https://universaldependencies.org/u/feat/Tense.html} We therefore leave TMA features to future work and instead only use agreement and transitivity tasks. We look at whether or not subtrees representing AVCs encode the same information about these tasks as FMVs. 
Note that there is a difference between transitivity as a lexical property of a verb and transitivity of a specific clause: some verbs can be used both transitively and intransitively \citep{aikhenvald00changing}. For practical reasons, we only consider transitivity as a property of a clause, rather than as a lexical property of a verb. \\
\indent An assumption underlying our question is that agreement and transitivity information is learned by the parser and specifically, that it is available to the parser when making decisions about main verbs. \citet{johnson2011relevant} found that adding subject-agreement features in the Charniak parser did not improve accuracy but explained this result by the fact that this information was already present in POS tags. This information is not present in POS tags in UD\footnote{And we are not using POS tags in our parser anyway.} and it seems reasonable to assume that agreement information is useful for parsing, which we however want to test. 

Our research questions can therefore be formulated as follows, with the first being a pre-condition for the others:

\begin{itemize}
  \item[\textbf{RQ1}] Is information about agreement and transitivity learned by the parser? 
  \item[\textbf{RQ2}] Does a sequential NN-based dependency parser learn the notion of dissociated nucleus?
  \item[\textbf{RQ3}] Does a dependency parser augmented with a recursive layer learn the notion of dissociated nucleus?
\end{itemize}

\noindent How we go about answering these questions will be explained in more detail in \cref{sec:rqs}.

\section{Experimental set-up}
\label{sec:exp}

\begin{table}[tbp]
    \centering
    \begin{tabular}{ll|ll|ll|ll}
        &    & \multicolumn{2}{l}{FMV} & \multicolumn{2}{l}{punct} & \multicolumn{2}{l}{AVC} \\
        &    & train       & dev       & train        & dev        & train        & dev       \\
        \toprule
        \multirow{4}{*}{T} & ca & 14K   & 2K  & 7K    & 964 & 12K   & 2K  \\
        & fi & 12K   & 1K  & 9K    & 1K  & 4K    & 458 \\
        & hr & 6K    & 803 & 4K    & 491 & 5K    & 653 \\
        & nl & 9K    & 618 & 6K    & 516 & 5K    & 251 \\
        \midrule
        \multirow{4}{*}{A} & ca & 14K   & 2K  & 7K    & 964 & 12K   & 2K  \\
        & fi & 10K   & 1K  & 8K    & 850 & 4K    & 443 \\
        & hr & 6K    & 803 & 4K    & 491 & 5K    & 653 \\
        & nl & 9K    & 618 & 6K    & 516 & 5K    & 246
    \end{tabular}
    \caption{Dataset sizes for the transitivity (T) and agreement (A) tasks for finite verbs and AVCs (FMV and AVC) as well as punctuation items.}
    \label{tab:data}
\end{table}

\subsection{Data}
\subsubsection*{Treebank Selection}
We select UD treebanks according to the criteria from \citet{delhoneux17old} that are relevant for this task:

\begin{itemize}
    \item Typological variety
    \item Variety of domains
    \item High annotation quality
\end{itemize}

\noindent We add criteria specific to our problem:
\begin{itemize}
    \item A minimum amount of AVCs (at least 4,000)
    \item High annotation quality of AVCs
    \item Availability of the information we need for the prediction tasks
\end{itemize}

\noindent
Tests of treebank quality are available on the UD homepage, and we can look at the quality of auxiliary chains to know if AVCs are well annotated.\footnote{http://universaldependencies.org/svalidation.html}
The information we need for the prediction tasks are: the presence of \emph{Verbform=Fin} in morphological features, so as to collect FMVs, and the presence of the Number and Person features for the agreement task.

These added criteria make the selection more difficult as it discards a lot of the small treebanks and makes it difficult to keep typological variety, since the bigger treebanks come from the same language families and we want to avoid having results that are biased in terms of typological properties. We select Catalan AnCora (ca), Croatian SET (hr), Dutch Alpino (nl) and Finnish TDT (fi). Table~\ref{tab:data} summarizes the data used. We use UD version 2.2 \citep{ud22}. 
Note that for Catalan, we use a list of lemmas from the UD documentation to filter out noisy cases of auxilary dependency relations which are numerous.

\subsubsection*{Creating the Dataset}
\begin{figure}[t]
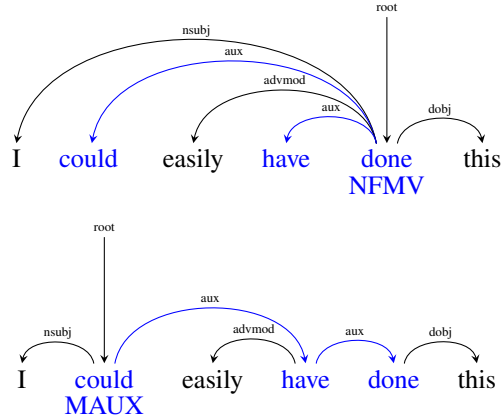

  \centering
  \begin{dependency}[theme = simple]
    \begin{deptext}[column sep=1em]
      I \& \textcolor{blue}{could } \& easily \& \textcolor{blue}{have} \& \textcolor{blue}{done} \& this \\
      \& \& \& \& \textcolor{blue}{NFMV} \& \\
    \end{deptext}
    \depedge{5}{1}{nsubj}
    \depedge[blue]{5}{2}{aux}
    \depedge{5}{3}{advmod}
    \depedge[blue]{5}{4}{aux}
    \deproot{5}{root}
    \depedge{5}{6}{dobj}
  \end{dependency}

  \begin{dependency}[theme = simple]
    \begin{deptext}[column sep=1em]
      I \& \textcolor{blue}{could} \& easily \& \textcolor{blue}{have} \& \textcolor{blue}{done} \& this \\
      \& \textcolor{blue}{MAUX} \& \& \& \& \\
    \end{deptext}
    \depedge{2}{1}{nsubj}
    \depedge{4}{3}{advmod}
    \depedge[blue]{2}{4}{aux}
    \depedge[blue]{4}{5}{aux}
    \deproot{2}{root}
    \depedge{5}{6}{dobj}
  \end{dependency}
  \caption{Example sentence with an AVC annotated in UD (top), and in MS (bottom). AVC subtree in thick blue.}
  \label{fig:avc_ud_ms}
\end{figure}

\indent We are mostly interested in what happens in parsing with the UD representation of AVCs since we believe it is a sound representation as argued in the introduction. This is because function words attach to content words which is compatible with an interpretation where these relations are part of a dissociated nucleus. However, it is also informative to look at what happens with a representation where auxiliaries are the head of AVCs. We do this in order to find out if the representation of the AVC subtree differs depending on which element is its head, since elements of the AVC share head properties. We therefore also consider the representation described in \citet{melcuk88} (Mel'\v{c}uk style, henceforth MS). We use the method in \citet{delhoneux16should} to transform the datasets from UD to MS.\footnote{With a slight modification: in that method, we discarded AVCs if they were in the passive voice, here we do not.} An example of AVC as represented in UD is given in the top part of Figure~\ref{fig:avc_ud_ms} and its transformed representation into MS is given in the bottom part of that figure.

\paragraph{Collecting FMVs \& AVCs in UD}

\begin{figure}[t]
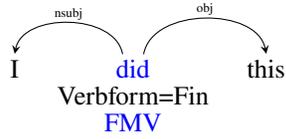

  \centering
  \begin{dependency}[theme = simple]
    \begin{deptext}[column sep=1em]
      I \& \textcolor{blue}{did} \& this \\
      \& Verbform=Fin \&\\
      \& \textcolor{blue}{FMV} \&\\
    \end{deptext}
    \depedge{2}{1}{nsubj}
    \depedge{2}{3}{obj}
  \end{dependency}
  \caption{Finite main verb in a UD tree.}
  \label{fig:fmv}
\end{figure}

\indent Collecting FMVs such as in Figure~\ref{fig:fmv} in UD treebanks is straightforward: verbs are annotated with a feature called \emph{VerbForm} which has the value \emph{Fin} if the verb is finite. We find candidates using the feature \emph{VerbForm=Fin} and only keep those that are not dependent of a copula or auxiliary dependency relations to make sure they are not part of a larger verbal construction.\\
\indent Collecting AVCs like the one in Figure~\ref{fig:avc_ud_ms} is slightly more involved. We do this in the same way as \citet{delhoneux16should}. 
In UD (top of the figure), an AVC $v_i$ has a main verb $mv_i$ (\emph{done} in the example) and a set of auxiliaries $AUX_i$ with at least one element (\emph{could} and \emph{have} in the example). AVCs are collected by traversing the sentence from left to right, looking at auxiliary dependency relations and collecting information about the AVCs that these relations are a part of. An auxiliary dependency relation $w_{aux}{\xleftarrow{aux}} w_{mv}$ is a relation where the main verb is the head and the auxiliary is the dependent (and the two verbs may occur in any order). 
Only auxiliary dependency relations between two verbal forms are considered. This allows us to filter out cases where a noun is head of an auxiliary dependency relation and making sure we have a main verb.  
We maintain a dictionary of AVC main verbs. When we find an auxiliary dependency relation, we add the dependent to the set of auxiliaries $AUX_i$ of the AVC whose main verb $mv_i$ is the head of that dependency relation.

\paragraph{Collecting AVCs in MS}
To collect AVCs in MS such as in the bottom of Figure~\ref{fig:avc_ud_ms}, we also scan the sentence left to right, looking for auxiliary dependency relations and we maintain a list of auxilaries which are part of AVCs of the sentence. When we find an auxiliary dependency relation, if its dependent is not in the list of auxiliaries already processed, we follow the chain of heads until we find an auxiliary which is not itself the dependent of an auxiliary relation. We then follow the chain of dependents until we find a node which is not the head of an auxiliary dependency relation, which is the main verb. While recursing the auxiliary chain, we add each head of an auxiliary dependency relation to the list of auxiliaries for the sentence.

\paragraph{Tasks}
When we have our set of FMVs and AVCs, we can create our task data sets. 
The transitivity task is a binary decision of whether the main verb has an object or not. This information can be obtained by looking at whether or not the main verb has an \emph{obj} dependent. In UD, a verb can have only one such dependent.\footnote{We experimented with a harder task: predicting the number of objects. In the example in Figure~\ref{fig:avc_ud_ms}, that number would be 1. In case of intransitive use of verbs, it would be 0, and with a ditransitive use of a verb, it would be 2. We observed the same trends and therefore do not report these results.}\\ 
\indent For the agreement task, we look at the morphological features of the verbs (the FMV or the auxiliary in case of AVCs) and concatenate the features \emph{Person} and \emph{Number}. The possible values are therefore all possible combination of 1st, 2nd and 3rd person with plural and singular. There are cases where this information is not available, in which case the agreement task is undefined for the AVC. \\
\indent The code to reproduce our experiments is available at \url{https://github.com/mdelhoneux/avc_analyser}, including the modifications we made to the parser to freeze the vector representations at the different layers in the network.

\begin{table}[tbp]
    \center
    \begin{tabular}{l|llll}
        & ud\_bas & ud\_rc & ms\_bas & ms\_rc \\
        \toprule
        ca & 87.8 & 87.7 & 87.8 & 87.8 \\
        fi & 78.9 & 78.9 & 78.6 & 78.7 \\
        hr & 81.0 & 80.8 & 80.5 & 80.6 \\
        nl & 84.1 & 83.7 & 84.1 & 83.7 \\
        \midrule
        av & 83.0 & 82.8 & 82.7 & 82.7 \\
    \end{tabular}
    \caption{LAS results of the baseline ($bas$) and recursive ($rec$) parser with UD and MS representations.\label{tab:parsing_res}}
\end{table}

\subsection{BiLSTM-based Parsing}
\label{sec:parser}

We use UUParser, a greedy transition-based parser \citep{nivre2008algorithms} based on the framework of \citet{kiperwasser16} where
BiLSTMs \citep{hochreiter1997long,graves2008bilstms} learn representations of tokens in context, and are trained together with a multi-layer perceptron that predicts transitions and arc labels based on a few Bi\-LSTM vectors.
Our parser uses the arc-hybrid transition system from \citet{kuhlmann11}
and is extended with a \textsc{Swap} transition to 
allow the construction of non-projective dependency trees \cite{nivre09acl}.
We also introduce a static-dynamic oracle to allow the parser to learn from non-optimal configurations at training time in order to recover better from mistakes at test time \citep{delhoneux17arc}.

\indent For an input sentence of length $n$ with words $w_1,\dots,w_n$, the parser creates a sequence of vectors $x_{1:n}$, where the vector $x_i$ representing $w_i$ is the concatenation of a randomly initialized word embedding $e(w_i)$ and a character vector. The character vector is obtained by running a \mbox{BiLSTM} over the characters $ch_j$ ($1 \leq j \leq m$) of $w_i$. Finally, each input element is represented by a \mbox{BiLSTM} vector, $v_i$:
\begin{align}
  x_i &= [e(w_i);\textsc{BiLstm}(ch_{1:m})]\\
  v_i &= \textsc{BiLstm}(x_{1:n},i)
\end{align}

\noindent The parser therefore learns representations at a type level which consists of two parts: 1) an embedding of the word type which represents its use in the corpus ($e(w_i)$) and 2) a character vector representing the sequence of characters of the word type. It also learns a representation of the word at the token level, in the context of the sentence ($v_i$). We will refer to these as \emph{type}, \emph{character} and \emph{token} vectors respectively.

\indent As is usual in transition-based parsing, the parser makes use of a configuration which consists of a stack, a buffer and a set of arcs. The configuration $c$ is represented by a feature function $\phi(\cdot)$ over a subset of its elements and for each configuration, transitions are scored by a classifier. In this case, the classifier is a multi-layer perceptron (MLP) and $\phi(\cdot)$ is a concatenation of \mbox{BiLSTM} vectors on top of the stack and the beginning of the buffer. The MLP scores transitions together with the arc labels for transitions that involve adding an arc. Both the embeddings and the \mbox{BiLSTMs} are trained together with the model. For simplicity, we only use the 2 top items of the stack and the first item of the buffer, as they are the tokens that may be involved in a transition in this transition system.

\paragraph*{AVC subtree vector}
As explained earlier, we are interested in finding out whether or not an LSTM trained with a parsing objective can learn the notion of dissociated nucleus as well as if a recursive composition function can help to learn this. The head of an AVC in UD is a non-finite main verb, which we will refer to as NFMV. The head of an AVC in MS is the outermost auxiliary, which we will refer to as the main auxiliary MAUX. We therefore look at NFMV and MAUX token vectors for the respective representation schemes and consider two definitions of these one where we use the BiLSTM encoding of the main verb token $v_i$. In the other, we construct a subtree vector $c_i$ by recursively composing the representation of AVCs as auxiliaries get attached to their main verb. When training the parser, we concatenate this composed vector to a vector of the head of the subtree to form $v_i$. This makes little difference in parsing accuracy, see Table~\ref{tab:parsing_res}.\\
\indent As in \citet{delhoneux19recursive}, we follow \citet{dyer15} in defining the composition function. The composed representation $c_i$ is built by concatenating the token vector $v_h$ of the head with the vector of the dependent $v_d$ being attached, as well as a vector $r$ representing the label used and the direction of the arc, see Equation~\ref{eq:compos}. (In our case, since we are only composing the subtrees of AVCs, $r$ can only have two values: left-aux and right-aux.) That concatenated vector is passed through an affine transformation and then a ($tanh$) non-linear activation. Initially, $c_i$ is just a copy of the token vector of the word ($\textsc{BiLstm}(x_{1:n},i)$).
\begin{align}
  v_i &= [\textsc{BiLstm}(x_{1:n},i) ; c_i]\\
  c_i &= tanh(W[v_h;v_d;r]+b) \label{eq:compos}
\end{align}
For our prediction experiments using this recursive composition function, we only investigate what is encoded in $c_i$. We call this experimental setup the \emph{recursive} setup. We refer to this recursive composition function as \emph{composition} in the remainder of this article.
Note that in \citet{delhoneux19recursive} we used two different composition functions, one using a simple recurrent cell and one using an LSTM cell. We saw that the one using an LSTM cell performed better. However, in this set of experiments, we only do recursive compositions over a limited part of the subtree: only between auxiliaries and NFMVs. This means that the LSTM would only pass through two states in most cases, and maximum 4.\footnote{The maximum number of auxiliaries in one AVC in our dataset is 3.} This does not allow us to learn proper weights for the input, output and forget gates. An RNN seems more appropriate here and we only use that. 

\subsection{Research Questions Revisited}
\label{sec:rqs}
\indent Now that we have introduced all experimental variables, we can explain in more detail than before how we can go about answering our main research questions, repeated below, and then explained in more detail in turn.
\begin{itemize}
  \item[\textbf{RQ1}] Is information about agreement and transitivity learned by the parser? 
  \item[\textbf{RQ2}] Does a sequential NN-based dependency parser learn the notion of dissociated nucleus?
  \item[\textbf{RQ3}] Does a dependency parser augmented with a recursive layer learn the notion of dissociated nucleus?
\end{itemize}

\paragraph*{RQ1}

\indent We verify that the parser has information about transitivity and agreement available when making parsing decisions about FMVs by comparing the accuracy of classifiers trained on token vectors of FMVs on these tasks to the majority baseline (the most frequent value for the task in training data). We expect them to perform substantially better than that.\\
\indent We also want to compare the representation of FMVs with the representation of a nearby token that is not expected to have this information, to rule out the possibility that the LSTM propagates information about AVCs to all the sentence tokens or at least the nearby ones. We select punctuation items that are close to the main verb for this purpose. Punctuation items attach to the main verb in UD and are frequent enough that we can expect many main verbs to have at least one as a dependent. We expect these vectors to be uninformative about the tasks.

If this information is available in token vectors of FMVs, we are also interested in finding out how this information was obtained from the network. If it is available in the context independent representation of words, i.e.~the type and character representation of the word, it may be propagated upwards from these representations to the token representation. Otherwise, we know that it is learned by the BiLSTM.\\
\indent If it is present in the context independent representation of words, there is some indication that this information is learned by the parser. To verify that it is learned specifically for the task of parsing, we compare type vectors of FMVs obtained using our parser to vectors trained using a language modelling objective. We train a word2vec \citep{mikolov13word2vec} language model on the same training set as for parsing.  We thus obtain vectors of the same dimension as our word type vectors and trained with the same data set but trained with a different objective.\footnote{Note that for this kind of experiment, training language models that learn contextual representations of words such as ELMo \citep{elmo} or BERT \citep{devlin19} and compare these representations to our token vectors would be more appropriate. However, these models are typically trained on very large datasets and it is unclear how well they perform when trained on just treebank data. We leave doing this to future work.} 
We expect the following to hold:
\begin{itemize}
  \item Main verb token vectors are informative with respect to transitivity and agreement: they perform better than the majority baseline on these tasks.
  \item Main verb token vectors are more informative than punctuation token vectors with respect to transitivity and agreement.
  \item Main verb type vectors trained with a parsing objective contain more information about transitivity and agreement than type vectors trained with a language modelling objective. 
\end{itemize}

\paragraph*{RQ2 and RQ3}
\noindent If the parser learns a notion of dissociated nucleus we expect to observe that AVC subtree vectors (i.e.\ the AVC's NFMV token vector or its composed version for UD, the AVC's MAUX token vector or its composed version for MS) contain a similar amount of information about agreement and transitivity as FMVs do. \\
\indent Note that we investigate whether the parser learns information about these tasks. We think it is reasonable to assume that if this information is learned by the parser, it is useful for the task of parsing, since the representations are obtained from a model which is trained end-to-end without explicit supervision about the prediction tasks. Recent research has cast some doubt on this: as discussed by \citet{belinkov18internal}, it is not impossible that the information is in the network activations but is not used by the network. In any case, we are not interested in improving the parser here but in finding out what it learns. 

\subsection{Vectors}
\indent We train parsers for 30 epochs for all these treebanks and pick the model of the best epoch based on LAS score on the development set. We report parsing results in Table~\ref{tab:parsing_res}. We train the parser with the same hyperparameters as in \citet{smith2018st} except for the character BiLSTM values: we set the character embedding size to 24 and the character BiLSTM output dimension to 50. This is a good compromise between efficiency and accuracy. The token vectors have a dimension of 250, type vectors 100 and character vectors 50. We load the parameters of the trained parser, run the BiLSTM through the training and development sentences, and collect the token vectors of NFMVs or MAUX, as well as the type and token vectors of finite verbs, from the training and development sets. We use the vectors collected from the training set to train the classifiers and we test the classifiers on the vectors of the development sets.\\
\indent As punctuation vectors, we take the token vectors of the punctuation items that are closest to the FMV. We look at children of the FMV that have \emph{punct} as dependency relation and take the closest one in linear order, first looking at the right children and then at the left ones.\\
\indent As for word2vec vectors, we use the Gensim \citep{gensim} implementation with default settings: using CBOW, a window of 5 and ignoring words with lower frequency than 5. We learn embeddings of the same dimension as our type vectors for comparability: 100.\\
\indent In the \emph{recursive} setup, we load the best model, pass the data through the BiLSTM and parse the sentences so as to obtain composed representations of AVCs. For that, we collect the final composed vectors of NFMVs or MAUX after prediction for each sentence.\\ 
\begin{figure}[]
    \centering
    \includegraphics[scale=0.6]{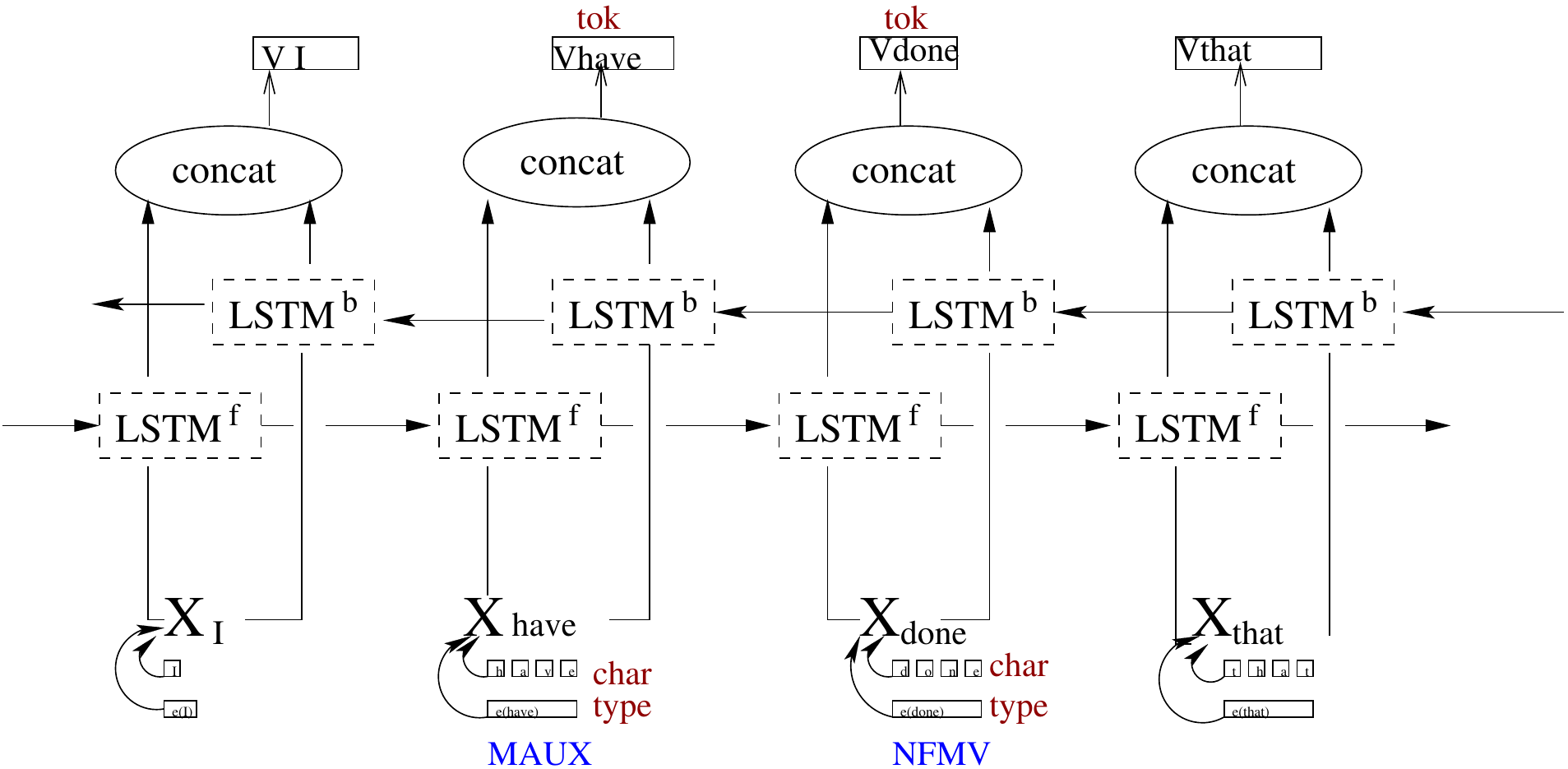}
    \caption{Example AVC with vectors of interest: token (tok), character (char) and type of MAUX and NFMV.}
    \label{fig:arch_avc}
\end{figure}
\begin{figure}[]
    \centering
    \includegraphics[scale=0.6]{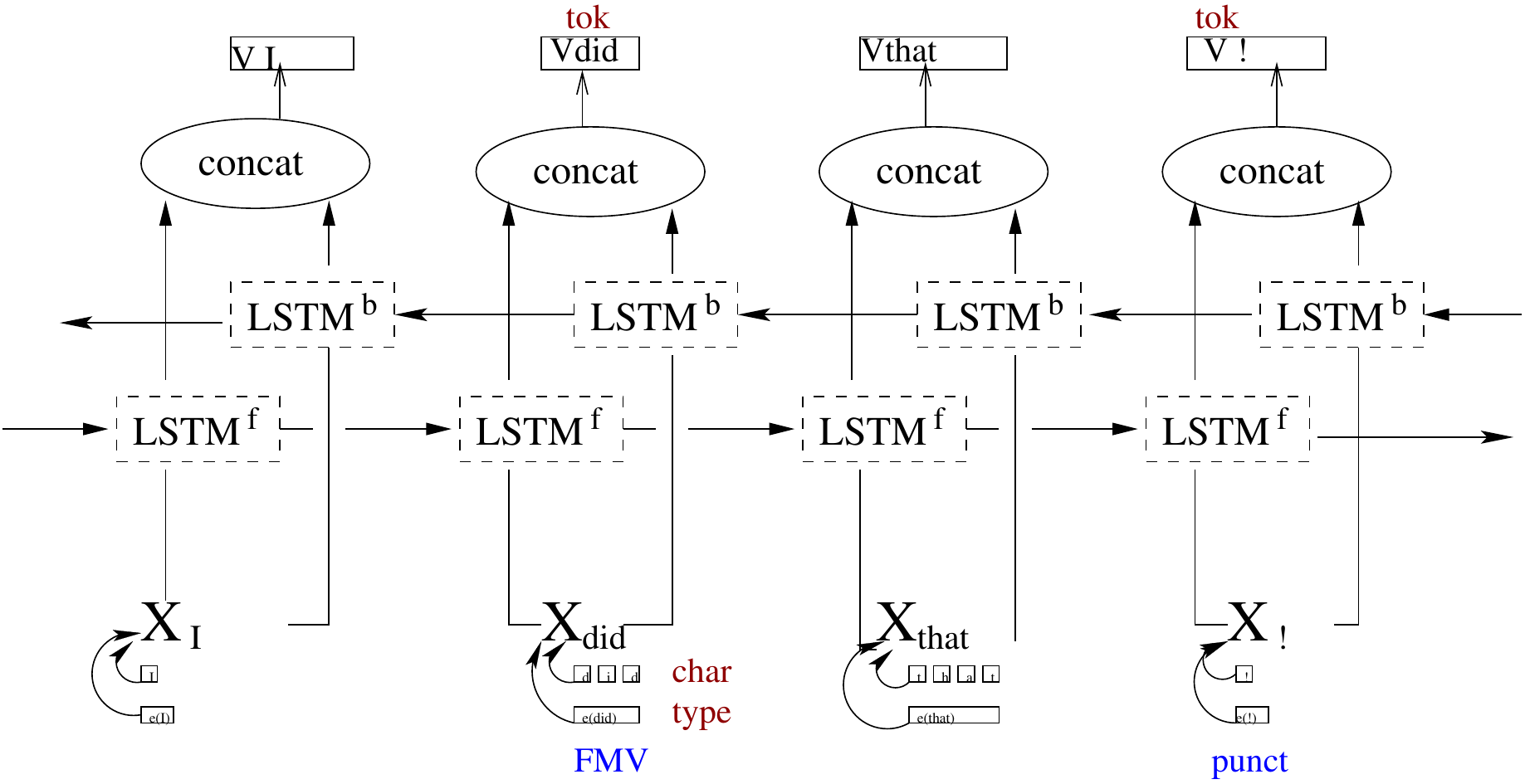}
    \caption{Example sentence with a FMV with vectors of interest: token (tok), character (char) and type of FMV and punctuation (punct).}
    \label{fig:arch_fmv}
\end{figure}
\indent We illustrate the vectors of interest with either representation in Figure~\ref{fig:arch_avc} and \ref{fig:arch_fmv}, except for the word2vec ones and the ones obtained recursively.

\indent We follow \citet{belinkov18internal} and choose to work with an MLP with one hidden layer as a diagnostic classifier. \citet{belinkov18internal} argues that it is important to avoid working with a classifier that is too weak or too strong. If it is too weak, it might fail to find information that is in the vectors investigated. If it is too strong, it might find patterns that the main network cannot use. He argues that a neural network with one hidden layer strikes a good balance in classifier power. In one study, he compared accuracies of a linear classifier and two non-linear classifiers with one and two hidden layers and found that while the non-linear classifiers performed consistently better on the task, the linear classifier showed the same trends. We ran all our experiments with a linear classifier as well and observed the same thing: the trends are generally the same. For the first research question, this means that the question can be answered positively regardless of the classifier used. For our other research questions, since they are more concerned with \emph{where} the information is present in the network than \emph{how much} information there is, this means that our observations do not depend on the classifier used and that our conclusions would be the same with different classifiers. For clarity, we therefore only report results using the MLP.

\section{Results}
\label{sec:res}

\renewcommand*{\MinNumber}{-20}%
\renewcommand*{\MidNumber}{0.0}%
\renewcommand*{\MaxNumber}{20}%

\begin{table}[tbp]
    \centering
    \setlength{\tabcolsep}{3.5pt}
    \begin{tabular}{ll|lllll|ll|RRRR|R}
        &       & \multicolumn{5}{l}{FMV}           & \multicolumn{2}{l}{punct} & \multicolumn{4}{l}{$\delta$ FMV} & $\delta$ punct \\
        &       & maj  & tok  & type & char  & w2v  & maj         & tok         & tok    & type   & char  & w2v  & tok     \EndTableHeader\\ 
        \toprule
        \multirow{6}{*}{ T } & ca & 70.5 & 88.7 & 79.4 & 75.0 & 74.4 & 67.5 & 71.3 & 18.2 & 9.0 & 4.5 & 3.9 & 3.8 \\
        & fi & 59.2 & 86.2 & 72.8 & 74.9 & 59.2 & 56.6 & 64.1 & 27.0 & 13.5 & 15.6 & 0.0 & 7.5 \\
        & hr & 55.9 & 79.7 & 71.3 & 70.6 & 57.8 & 61.5 & 62.7 & 23.8 & 15.4 & 14.7 & 1.8 & 1.2 \\
        & nl & 61.7 & 82.1 & 74.0 & 69.4 & 64.8 & 62.0 & 69.6 & 20.5 & 12.4 & 7.7 & 3.1 & 7.6 \\
        \midrule
        \StartTableHeader
        & av & 61.8 & 84.2 & 74.4 & 72.5 & 64.0 & 61.9 & 66.9 & 22.4** & 12.6* & 10.7* & 2.2 & 5.0* \\
        & sd & 6.2 & 4.0 & 3.5 & 2.9 & 7.5 & 4.5 & 4.2 & 3.9 & 2.7 & 5.4 & 1.7 & 3.1 \EndTableHeader\\
        \midrule
        \multirow{6}{*}{ A } & ca & 74.4 & 82.2 & 82.6 & 98.4 & 74.4 & 76.7 & 76.6 & 7.7 & 8.1 & 24.0 & 0.0 & -0.1 \\
        & fi & 61.6 & 86.0 & 63.2 & 93.5 & 61.6 & 59.7 & 59.7 & 24.4 & 1.6 & 31.9 & 0.0 & 0.0 \\
        & hr & 60.9 & 78.1 & 74.8 & 97.8 & 60.9 & 64.0 & 64.0 & 17.2 & 13.9 & 36.9 & 0.0 & 0.0 \\
        & nl & 81.6 & 87.2 & 85.7 & 96.3 & 81.6 & 81.8 & 80.5 & 5.7 & 4.2 & 14.8 & 0.0 & -1.2 \\
        \midrule
        \StartTableHeader
        & av & 69.6 & 83.4 & 76.6 & 96.5 & 69.6 & 70.5 & 70.2 & 13.7* & 6.9* & 26.9** & 0.0 & -0.3 \\
        & sd & 10.1 & 4.1 & 10.0 & 2.2 & 10.1 & 10.4 & 10.0 & 8.7 & 5.4 & 9.7 & 0.0 & 0.6 \EndTableHeader\\ 
    \end{tabular}
    \caption{Classification accuracy of the majority baseline (maj) and classifier trained on the type, token (tok) and word2vec (w2v) vectors of FMVs, and token vectors of punctuation (punct) on the agreement (A) and transitivity (T) tasks. Difference ($\delta$) to majority baseline of these classifiers. Average difference significantly higher than the baseline are marked with $* = p <.05$ and $** = p <.01$. }
    \label{tab:res1}
\end{table}

\indent We compare the prediction accuracy on each task to the majority baseline. To get a measure that is comparable across languages and settings, we compute the difference between the accuracy of a classifier on a task using a set of vectors to the majority baseline for this set of vectors. The larger the difference, the greater the indication that the vector encodes the information needed for the task.\footnote{We calculated relative error reductions as well but since these results showed the same trends, we exclusively report absolute difference in accuracy.} We perform paired t-tests to measure whether the difference to the majority baseline is statistically significant on average. Given that the set of labels we predict is very restricted, the majority baseline performs reasonably well. As can be seen in Table~\ref{tab:res1} and Table~\ref{tab:mvfv_res}, the performance ranges from 49 to 81 with most values around 60. It seems reasonable to assume that a system which performs significantly better than that learns information relevant to the notion predicted. The majority baselines of FMVs and AVCs are on average very close (see Table~\ref{tab:mvfv_res}), indicating that results are comparable for these two sets.\footnote{The colour scheme in the tables indicates no change (yellow) to improvements (dark green).}

\subsection*{RQ1: Is agreement and transitivity information learned by the parser?}
\indent Results pertaining to \textbf{RQ1} are given in Table~\ref{tab:res1}. Note that we only report results on the UD data here since the representation of finite verbs does not change between UD and MS. The results conform to our expectations. FMV token vectors contain information about both agreement and transitivity. They perform significantly better than the majority baseline for both tasks. \\
\indent Token vectors of punctuation marks related to FMVs contain some information about transitivity, with some variance depending on the language and perform significantly better than the majority baseline. However, they perform considerably worse than token vectors of FMVs. The difference between FMV token vectors and their majority baseline is substantially larger than the difference between punctuation vectors and their majority baseline for all languages and with 17 percentage points more on average. Punctuation vectors seem to be completely uninformative about agreement. The classifier seems to learn the majority baseline in most cases. This indicates that information about agreement and transitivity is relevant for scoring transitions involving FMVs but this information is not necessary to score transitions involving tokens that are related to FMVs (or at least it is not necessary for punctuation tokens that are related to FMVs). It indicates that this information is available in contextual information of FMVs but not in contextual information of all tokens in the verb phrase, except marginally for transitivity.\\
\indent We can conclude that information about both agreement and transitivity are learned by the parser. We can now look more closely at where in the network this information is present by looking at context independent vectors: type and character vectors. FMV type vectors seem to encode some information about transitivity: on average, they perform significantly better than the majority baseline on the transitivity task. When it comes to agreement, the difference to the baseline for FMV type vectors is smaller but they still perform significantly better than the majority baseline on average, although with more variation: they seem uninformative for Finnish. 
FMV token vectors contain substantially more information than type vectors for both tasks. The difference between token vectors and their majority baseline is, in most cases, slightly to substantially larger than it is between type vectors and their majority baseline, but on average, it is substantially larger. \\
\indent FMV token vectors also contain substantially more information than character vectors for the transitivity task. For the agreement task, however, character vectors contain substantially more information than token vectors.\\
\indent This indicates that the information flow for the two tasks differs to some extent. 
Since token vectors contain much more information about transitivity than both type and character vectors, we can conclude that this information is obtained from the BiLSTM.\footnote{Further evidence for this conclusion comes from the fact that some of this information, unlike the agreement information, seems to spill over on neighboring tokens like the punctuation tokens considered earlier.} For agreement, however, token vectors are less informative than character vectors which indicates that part of this information probably comes from the character vector, but some of it gets filtered out between the character vector and the token vector. This indicates that agreement may be useful for signalling potential relationships between words which are then captured by the BiLSTM. A substantial part of the information does remain though, indicating that agreement information is still useful when it comes to making parsing decisions. \\ 
\indent We finally compare representations of word types when trained for the parsing as opposed to the language modelling task. For this, we look at word2vec vectors. Word2vec vectors of FMVs contain little information about transitivity, from no difference with the majority baseline to 3.9 percentage points above it. FMV type vectors are substantially better than word2vec vectors for all languages, with an average difference to the majority baseline that is 10 percentage points larger than the difference between word2vec vectors and their majority baseline. Word2vec vectors contain no information at all about agreement, the network learns the majority baseline for these vectors for all languages. FMV type vectors are better on this task for all languages. The difference between FMV type vectors and the majority baseline is small for Finnish but on average it is 6.9 percentage points larger than the majority baseline. This indicates that information about transitivity and agreement is more relevant for the task of parsing than for the task of language modelling.\\
\indent We have clearly seen 1) that transitivity and agreement are learned by the parser and that some information about these tasks is available to the parser when making decisions about FMVs and 2) that this information is not available everywhere in the network and is therefore available specifically when making decisions about FMVs. This answers \textbf{RQ1} positively.\\
\indent We should keep in mind that we observed a different information flow for transitivity where information is obtained mostly by the BiLSTM compared to agreement where it seems to be strongly signalled at the layer of context independent representations (in particular in the character vector) and weaker at the output of the BiLSTM. 

\begin{table}[tbp]
    \centering
    \small
    \setlength{\tabcolsep}{1.5pt}
    \begin{tabular}{ll|ll|l|lll|lll|R|RRR|RRR}
        & & \multicolumn{2}{c}{ FMV } & AVC & \multicolumn{3}{c}{ NFMV-UD } & \multicolumn{3}{c}{ MAUX-MS } & FMV & \multicolumn{3}{c}{ $\delta$ nfmv } & \multicolumn{3}{c}{ $\delta$ maux } \\
        & & maj & tok & maj & tok & type & char & tok & type & char & tok & tok & type & char & tok & type & char \EndTableHeader\\ 
        \toprule
        \multirow{6}{*}{ T } & ca & 70.5 & 88.7 & 66.9 & 89.3 & 78.8 & 74.3 & 88.5 & 66.8 & 66.7 & 18.2 & 22.4 & 11.9 & 7.4 & 21.7 & -0.1 & -0.2 \\
        & fi & 59.2 & 86.2 & 49.1 & 81.8 & 70.7 & 71.1 & 72.4 & 62.0 & 61.0 & 27.0 & 32.7 & 21.5 & 21.9 & 23.3 & 12.8 & 11.8 \\
        & hr & 55.9 & 79.7 & 51.2 & 82.5 & 75.4 & 69.7 & 74.9 & 55.8 & 56.1 & 23.8 & 31.4 & 24.3 & 18.5 & 23.7 & 4.7 & 4.9 \\
        & nl & 61.7 & 82.1 & 70.5 & 88.6 & 74.5 & 71.5 & 86.8 & 80.2 & 81.6 & 20.5 & 18.1 & 4.0 & 1.0 & 16.3 & 9.6 & 11.1 \\
        \midrule
        \StartTableHeader
        & av & 61.8 & 84.2 & 59.4 & 85.6 & 74.8 & 71.6 & 80.6 & 66.2 & 66.3 & 22.4** & 26.1** & 15.4* & 12.2* & 21.2** & 6.8* & 6.9* \\
        & sd & 6.2 & 4.0 & 10.8 & 3.9 & 3.3 & 1.9 & 8.2 & 10.3 & 11.1 & 3.9 & 7.0 & 9.3 & 9.7 & 3.4 & 5.7 & 5.7 \EndTableHeader \\
        \midrule
        \multirow{6}{*}{ A } & ca & 74.4 & 82.2 & 76.0 & 76.0 & 76.0 & 76.0 & 83.3 & 99.3 & 99.7 & 7.7 & 0.0 & 0.0 & 0.0 & 7.2 & 23.2 & 23.7 \\
        & fi & 61.6 & 86.0 & 67.5 & 68.5 & 67.5 & 69.8 & 77.7 & 89.6 & 92.3 & 24.4 & 1.0 & 0.0 & 2.3 & 10.2 & 22.1 & 24.8 \\
        & hr & 60.9 & 78.1 & 71.2 & 71.2 & 71.9 & 71.2 & 77.2 & 95.3 & 94.6 & 17.2 & 0.0 & 0.7 & 0.0 & 6.0 & 24.0 & 23.4 \\
        & nl & 81.6 & 87.2 & 72.4 & 70.4 & 72.4 & 72.4 & 89.4 & 99.5 & 100.0 & 5.7 & -2.0 & 0.1 & 0.0 & 17.1 & 27.2 & 27.6 \\
        \midrule
        \StartTableHeader
        & av & 69.6 & 83.4 & 71.8 & 71.5 & 72.0 & 72.3 & 81.9 & 95.9 & 96.7 & 13.7* & -0.2 & 0.2 & 0.6 & 10.1* & 24.1** & 24.9** \\
        & sd & 10.1 & 4.1 & 3.5 & 3.2 & 3.5 & 2.7 & 5.7 & 4.6 & 3.8 & 8.7 & 1.2 & 0.3 & 1.2 & 5.0 & 2.2 & 1.9 \EndTableHeader \\
    \end{tabular}
    \caption{Classification accuracy of the majority baseline (maj) and classifier trained on the token (tok), type and character (char) vectors of FMVs, NFMVs and MAUX on agreement (A) and transitivity (T) tasks. Difference ($\delta$) to majority baseline of these classifiers. Average difference significantly higher than the baseline are marked with $* = p <.05$ and $** = p <.01$. }
    \label{tab:mvfv_res}
\end{table}

\subsection*{RQ2: Does a BiLSTM-based parser learn the notion of dissociated nucleus?}
\indent Results pertaining to \textbf{RQ2} are given in Table~\ref{tab:mvfv_res}. Comparing first FMV and NFMV token vectors, we can see that NFMVs are somewhat better than FMVs at the transitivity task but both perform substantially better than the majority baseline. On the agreement task, however, FMVs vectors perform substantially better than NFMV vectors. NFMV vectors seem completely uninformative when it comes to agreement, performing on average slightly worse (-0.2 percentage point) than the majority baseline. FMV vectors perform moderately (Dutch) to largely better (Finnish) than this depending on the language, with an average difference to the majority baseline of 13.7.\\
\indent An unpaired t-test reveals that the results of FMVs and NFMVs on the agreement task are significantly different with $p<.01$, which further supports the hypothesis that they do not capture the same information. The results between FMVs and NFMVs are not significantly different on the transitivity task. \\
\indent Looking at MAUX token vectors in an MS representation, we see that they are also substantially better than the majority baseline on the transitivity task, performing slightly worse than FMVs, on average. Contrary to NFMV token vectors, they perform significantly better than the majority baseline on the agreement task and somewhat worse than FMVs. The results between MAUX and FMVs are
not significantly different for either of the tasks, indicating that they do seem to capture a similar amount of information.\\
\indent We can conclude that a BiLSTM-based parser does not learn the notion of dissociated nucleus for AVCs when working with a representation of AVCs where the main verb is the head such as is the case in UD: the representation of NFMVs contains less information about agreement than the representation of FMVs. However, when using a representation where the auxiliary is the head, a BiLSTM-based parser does seem to learn this notion, it learns a representation of the AVC's head that is similar to the representation of FMVs.\\ 
\begin{figure}[t]
    \centering
    \includegraphics[scale=0.6]{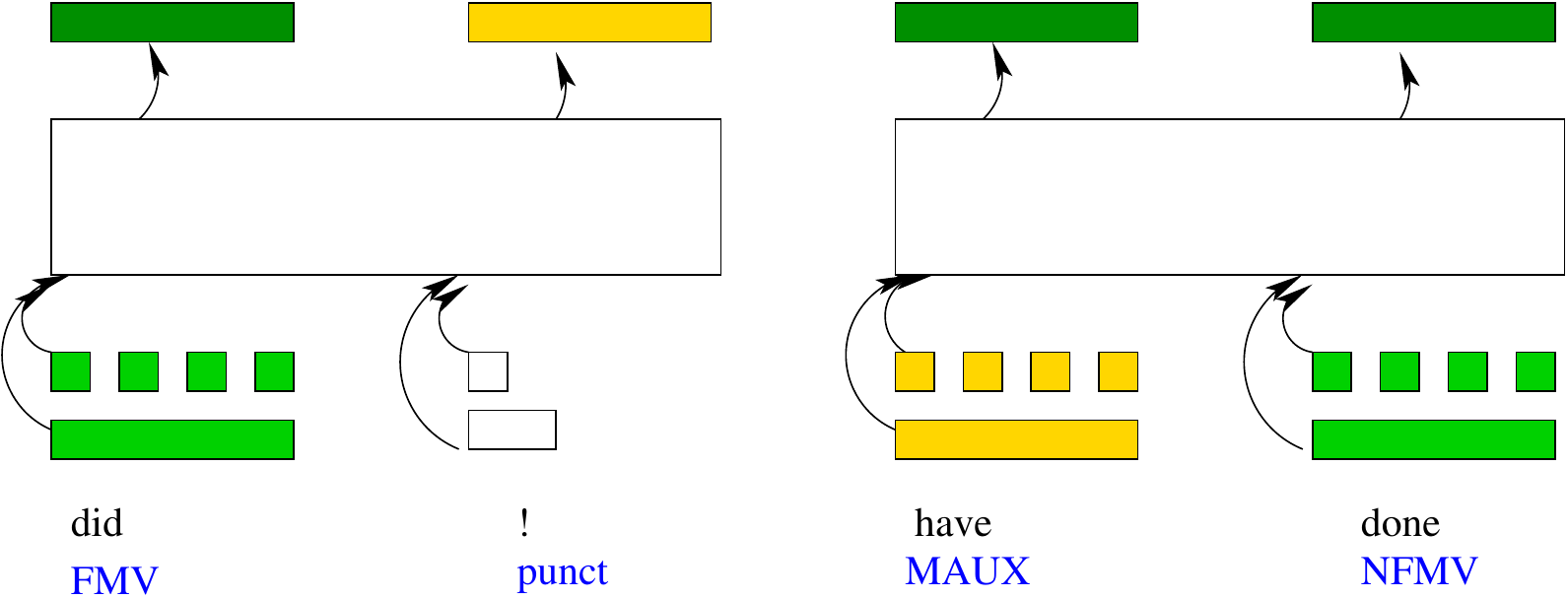}
    \caption{Information flow for transitivity.}
    \label{fig:avc_t}
\end{figure}
\begin{figure}[t]
    \centering
    \includegraphics[scale=0.6]{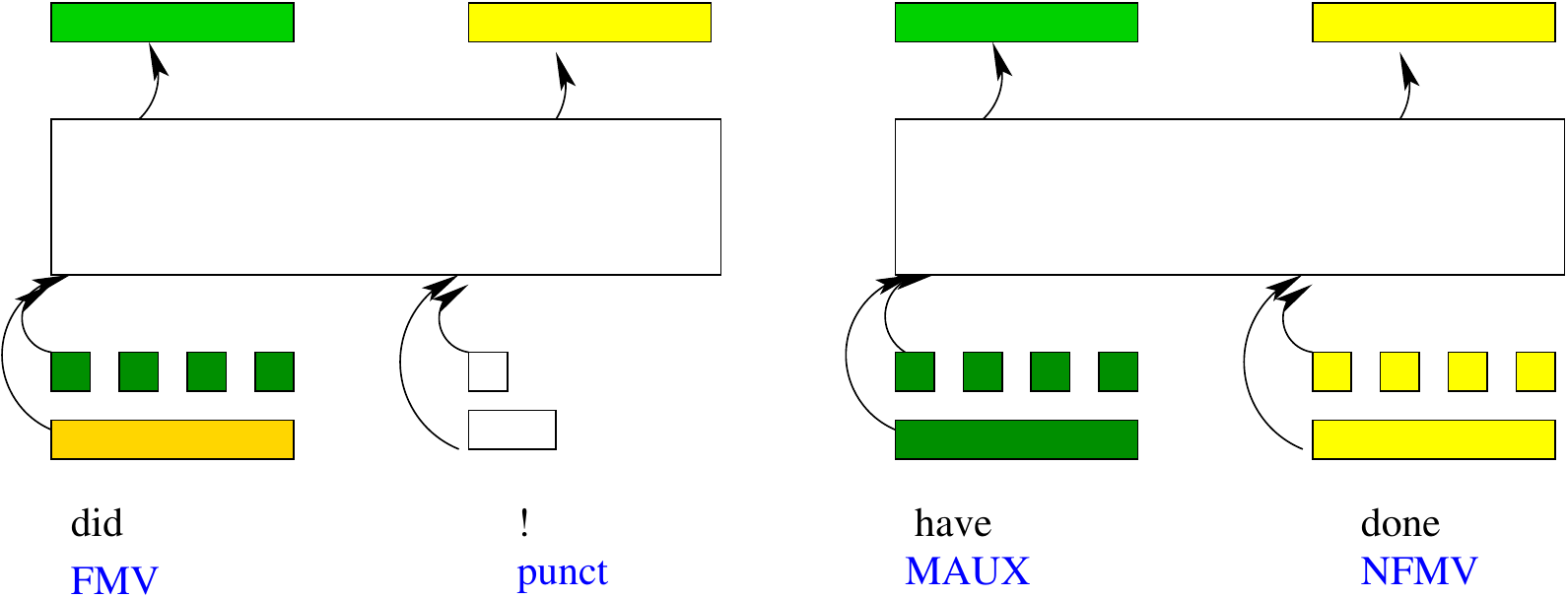}
    \caption{Information flow for agreement.}
    \label{fig:avc_a}
\end{figure}
\indent This can be explained by the different information flow in the network for the two tasks. In Figures~\ref{fig:avc_t} and~\ref{fig:avc_a}, we illustrate further the different information flow for the transitivity and agreement task respectively and for both FMVs and AVCs. We use the same colour scheme as in the tables (from yellow to dark green means no information about a task to a substantial amount of information about the task, as measured by the difference to the majority baseline) and we simplify the architecture illustration from Figure~\ref{fig:arch_avc} and~\ref{fig:arch_fmv}. As we saw in Table~\ref{tab:res1} and as we can see from the left part of these figures, looking at FMVs, information about transitivity is mostly obtained from the BiLSTM whereas information about agreement is present in the character vector and propagated to the token vector. 
We observe a similar phenomenon with AVCs, as presented in Table~\ref{tab:mvfv_res}: for the transitivity task, the type vectors of NFMVs contain more information than the type vectors of MAUX but in both cases, the token representation of the head of the AVC contains substantially more information than the type and character vectors. By contrast, both type and character vectors of MAUX contain information about agreement, whereas NFMV type and character vectors do not.
It seems that, with this model, in order for agreement information to be available to the head of the AVC, the head of the AVC needs to be the auxiliary. When it comes to transitivity, the BiLSTM is able to construct this information regardless of what word is the head of the AVC, which is why the model is able to learn the notion of dissociated nucleus for the MS representation but not the UD representation.\\
\indent Since the MS representation does not improve parsing accuracy (see Table~\ref{tab:parsing_res}\footnote{Note that these results are not directly comparable, we would need to transform the MS representation back to UD to compare against the same annotation type. However, we expect these results to be even worse when transformed, as results from \citet{delhoneux16should} indicate.}), it is possible that either learning this notion is not important for parsing, or that the benefits of learning this notion are offset by other factors. We attempt to find out whether or not we can get the best of both worlds by learning a BiLSTM parser that uses recursive composition and train it on the UD representation. We also look at what happens with recursive composition with the MS representation.\\
\indent Note that, in all experiments, we only report results for NFMVs with a UD representation and for MAUX with a MS representation because these are the vectors that represent the subtree. However, the information that is learned in these vectors does not seem to depend much on the representation of the AVC: token, type and character vectors representing MAUX learn similar information whether in UD or in MS and token, type and character vectors of NFMVs also learn similar information whether in UD or in MS. This means that the parser learns similar representations of AVC elements and only the representation of the subtree depends on the representation style of AVCs.

\begin{table}[tbp]
    \centering
    \setlength{\tabcolsep}{2.5pt}
    \begin{tabular}{l|l|ll|lllll|RRRRR}
        & & \multicolumn{2}{c}{ FMV } & AVC & \multicolumn{2}{c}{ NFMV } & \multicolumn{2}{c}{ MAUX } & \multicolumn{5}{c}{ $\delta$ } \\
        & & maj & tok & maj & tok & tok+c & tok & tok+c & fmv & nfmv & nfmv+c & maux & maux+c \EndTableHeader\\
        \toprule
        \multirow{6}{*}{ T } & ca & 70.5 & 88.7 & 66.9 & 89.3 & 88.5 & 88.5 & 86.5 & 18.2 & 22.4 & 21.7 & 21.7 & 19.7 \\
        & fi & 59.2 & 86.2 & 49.1 & 81.8 & 76.0 & 72.4 & 79.4 & 27.0 & 32.7 & 26.9 & 23.3 & 30.3 \\
        & hr & 55.9 & 79.7 & 51.2 & 82.5 & 82.0 & 74.9 & 79.5 & 23.8 & 31.4 & 30.9 & 23.7 & 28.3 \\
        & nl & 61.7 & 82.1 & 70.5 & 88.6 & 83.4 & 86.8 & 88.5 & 20.5 & 18.1 & 12.9 & 16.3 & 17.9 \\
        \midrule
        \StartTableHeader
        & av & 61.8 & 84.2 & 59.4 & 85.6 & 82.5 & 80.6 & 83.5 & 22.4** & 26.1** & 23.1** & 21.2** & 24.0** \\
        & sd & 6.2 & 4.0 & 10.8 & 3.9 & 5.2 & 8.2 & 4.7 & 3.9 & 7.0 & 7.8 & 3.4 & 6.2 \EndTableHeader\\
        \midrule
        \multirow{6}{*}{ A } & ca & 74.4 & 82.2 & 76.0 & 76.0 & 91.6 & 83.3 & 77.1 & 7.7 & 0.0 & 15.6 & 7.2 & 1.1 \\
        & fi & 61.6 & 86.0 & 67.5 & 68.5 & 79.2 & 77.7 & 74.7 & 24.4 & 1.0 & 11.7 & 10.2 & 7.2 \\
        & hr & 60.9 & 78.1 & 71.2 & 71.2 & 83.6 & 77.2 & 73.3 & 17.2 & 0.0 & 12.4 & 6.0 & 2.1 \\
        & nl & 81.6 & 87.2 & 72.4 & 70.4 & 84.0 & 89.4 & 83.6 & 5.7 & -2.0 & 11.6 & 17.1 & 11.2 \\
        \midrule
        \StartTableHeader
        & av & 69.6 & 83.4 & 71.8 & 71.5 & 84.6 & 81.9 & 77.2 & 13.7* & -0.2 & 12.8* & 10.1* & 5.4 \\
        & sd & 10.1 & 4.1 & 3.5 & 3.2 & 5.2 & 5.7 & 4.6 & 8.7 & 1.2 & 1.9 & 5.0 & 4.7 \EndTableHeader\\
    \end{tabular}
    \caption{Classification accuracy of the majority baseline (maj) and classifier trained on the vectors (vec) of NFMVs with ($+c$) and without composition and FMVs on agreement (A) and transitivity (T) tasks. Difference ($\delta$) to majority baseline of these classifiers. Average difference significantly higher than the baseline are marked with $* = p <.05$ and $** = p <.01$. }
    \label{tab:compos}
\end{table}

\subsection*{RQ3: Does subtree composition help?}
\indent As mentioned previously, we found in \citet{delhoneux19recursive} that a recursive composition function does not make our parsing model more accurate. A recursive composition function might not be necessary for parsing accuracy but might help in this case, however. It could make it possible to get the relevant information from the main verb and the auxiliary token vectors. As we have just seen, the token vector of the MAUX has information that is missing in the token vector of the NFMV and that could be propagated to the NFMV vector through the recursive composition function.\\
\indent We train a version of the parser where we recursively compose the representation of AVC subtrees during parsing. For UD, this means that the representation of the NFMV token gets updated as auxiliaries get attached to it. For MS, this means that the representation of AVC subtrees is composed in a chain from the outermost auxiliary to the main verb.
Note that this recursive composition function models the transfer relation for UD: it is used when an auxiliary is attached to the main verb. In MS, by contrast, it can also be used between two auxiliaries, which is a different type of relation. As reported in Table~\ref{tab:parsing_res}, this decreases parsing accuracy very slightly.
We compare the vectors of this composed representation to the representation of FMVs. Results are given in Table~\ref{tab:compos}.

\indent On average, composed NFMV vectors perform similarly to non-composed NFMV vectors on the transitivity task, slightly worse but substantially better than the majority baseline. For the agreement task, composed NFMV vectors are much better than the non-composed NFMV vectors, all performing substantially better than the majority baseline, although with less variation than the FMV vectors. 
The difference between composed NFMV vectors and the majority baseline is slightly higher (0.7 percentage points) than the difference between FMV vectors and their majority baseline for transitivity and slightly lower for agreement, but with variation across languages. On average, they seem to capture a similar amount of information. 
An unpaired t-test reveals that there is no significant difference between the results of FMVs and composed NFMV vectors. We can therefore conclude that a recursive composition function on top of a BiLSTM allows the model to capture similar information about AVCs and their non-dissociated counterpart, FMVs. This indicates that composing subtree representations with a recursive layer makes it possible for the parser to learn the notion of dissociated nucleus with a representation of AVCs where the head is the main verb.\\
\indent With the MS representation, composition improves accuracy on transitivity (except for Catalan) but decreases accuracy on agreement, making it on average not significantly better than the majority baseline and making the MAUX representation almost completely uninformative with regards to agreement for Catalan and Croatian. There is however no statistical difference between the results of FMVs and composed MAUX on either of the tasks, indicating that they do capture a similar amount of information.\\
\indent Overall, it seems that using a UD representation and a recursive composition function is the best option we have to have an accurate parser which captures the notion of dissociated nucleus, given our definition of what it means to capture this notion (that FMV and NFMV representations encode a similar amount of information about agreement and transitivity). This does not improve overall parsing accuracy which means either that it is not important to capture this notion for parsing or that the benefits of doing so are offset by drawbacks of this method. It would be interesting to find out whether learning this information is important to downstream tasks, which we leave to future work.

\section{Conclusion}
\indent We used diagnostic classifiers to investigate the question of whether or not a BiLSTM-based parser learns the notion of dissociated nucleus, focusing on AVCs. We looked at agreement and transitivity tasks and verified that the parser has access to this information when making parsing decisions concerning main verbs. We compared what a parser learns about AVCs with what it learns about their non-dissociated counterpart: finite main verbs. We observed that, with a UD representation of AVCs, vectors that represent AVCs encode information about transitivity to the same extent as FMVs but, contrary to FMVs, they are mostly uninformative when it comes to agreement. We concluded that a purely recurrent BiLSTM-based parser does not learn the notion of dissociated nucleus. We found explanations for this by investigating the information flow in the network and looking at what happens with a representation of AVCs where the auxiliary is the head.\\
\indent We finally investigated whether or not explicitly composing AVC subtree representations using a recursive layer makes a difference and it seems to make the representation of AVCs more similar to the representation of FMVs, indicating that recursively composing the representation of subtrees makes it possible for the parser to learn the notion of dissociated nucleus.\\
\indent We started out by arguing that parsers \emph{should} learn the notion of dissociated nucleus, then found out that a BiLSTM-based parser \emph{does} not learn this notion when working with a UD representation of AVCs but \emph{can} learn it if augmented by a recursive composition function. This recursive layer has been shown to be superfluous in previous work when we only look at parsing accuracy but our results here indicate that there may be benefits to using this recursive layer that are not reflected in parsing accuracy. This suggests that we may just not have found the best way to integrate this recursive layer into a BiLSTM-based parser yet. More generally, this lends some support to the hypothesis that hierarchical modelling is a useful inductive bias when modelling syntax. \\
\indent In future work, we plan to use diagnostic classifiers to investigate other cases of dissociated nuclei, such as combinations of adpositions and nouns and of subordinating conjunctions and verbs. We also plan to further investigate the use of a recursive layer in parsing with BiLSTM-based feature representations. 
It may be useful to model the \emph{transfer} relation between two elements of a dissociated nucleus to obtain a representation of the nucleus which is embedded in the same space as other nuclei. It would finally be interesting to find out whether learning this notion of dissociated nucleus is useful when it comes to downstream applications.

\starttwocolumn
\section*{Acknowledgments}
We thank Paola Merlo for thorough comments on an earlier version of this paper, as well as the anonymous reviewers. We acknowledge the computational resources provided by CSC in Helsinki and Sigma2 in Oslo through NeIC-NLPL (www.nlpl.eu).

\bibliography{main}

\end{document}